\documentclass[letterpaper, 10 pt, conference]{ieeeconf}  %

\IEEEoverridecommandlockouts                              %

\usepackage{graphics} %
\usepackage{graphicx}

\usepackage[caption=false]{subfig}
\usepackage{epsfig} %
\usepackage{amsmath} %
\usepackage{amssymb}  %
\usepackage{amsfonts}
\usepackage{bm}

\usepackage{algorithm}
\usepackage[noend]{algpseudocode} %
\algnewcommand\AAND{\textbf{ and }}
\algnewcommand\Or{\textbf{ or }}

\usepackage{color}
\usepackage{citesort}
\usepackage{flushend}
\usepackage{url}
\usepackage[breaklinks]{hyperref}
\usepackage[capitalise]{cleveref}
\usepackage{breakurl}
\usepackage{booktabs}
\usepackage{makecell}
\usepackage{multirow}

\usepackage[nolist,nohyperlinks]{acronym}
\acrodef{method}[AOM]{ACRONYM OF METHOD}
\acrodef{mc}[MC]{Microcontroller}
\acrodef{sbc}[SBC]{Single Board Computer}
\acrodef{obc}[OBC]{On-Board Computer}
\acrodef{rtc}[RTC]{Real-Time Clock}
\acrodef{pps}[PPS]{Pulse Per Second}
\acrodef{gnss}[GNSS]{Global Navigation Satellite Systems}
\acrodef{tov}[ToV]{Time of Validity}
\acrodef{imu}[IMU]{Inertial Measurement Unit}
\acrodef{ros}[ROS]{Robot Operating System}
\acrodef{pi}[PI]{Proportional-Integral}
\acrodef{ptp}[PTP]{Precision Time Protocal}

\usepackage{siunitx}
\usepackage{lipsum}

\title{\LARGE \bf
Simultaneous Triggering and Synchronization of Sensors and Onboard Computers
}

\author{Morten Nissov, Nikhil Khedekar, and Kostas Alexis%
\thanks{This material was supported by the Research Council of Norway Award NO-321435.}%
\thanks{The authors are with the Norwegian University of Science and Technology (NTNU), O. S. Bragstads Plass 2D, 7034, Trondheim, Norway {\tt\small morten.nissov@ntnu.no}}%
}

\begin{document}

\maketitle
\thispagestyle{empty}
\pagestyle{empty}

\begin{abstract}
High fidelity estimation algorithms for robotics require accurate data. However, timestamping of sensor data is a key issue that rarely receives the attention it deserves. Inaccurate timestamping can be compensated for in post-processing but is imperative for online estimation. Simultaneously, even online mitigation of timing issues can be achieved through a relaxation of the tuning parameters from their otherwise more performative optimal values, but at a detriment to performance.
To address the need for real-time, low-cost timestamping, a versatile system which utilizes readily-available components and established methods for synchronization is introduced. The synchronization
and triggering (of both high- and low-rate sensors) capabilities of the system are demonstrated.
\end{abstract}

\section{INTRODUCTION}\label{sec:introduction}
Accurately timestamped data is a necessity for building accurate state estimation algorithms, especially in the case of Simultaneous Localization And Mapping (SLAM) at high speeds. Although the research community has worked on developing methods that account for missing hardware synchronization, solving the problem at the root and producing high-accuracy synchronized data from the start greatly simplifies downstream complexity. Previous methods for synchronization have been attempted however, many of such methods~\cite{nikolic2014_vi_sensor,tschopp2020versavis} specialize to very specific sensor configurations. Furthermore, even the more general designs, able to synchronize a greater variety of sensors, focus only on the data gathering problem~\cite{albrektsen2017sync}. This leaves the main computer running on an unsynchronized clock, thereby greatly complicating closed-loop performance as synchronization over serial is limited by large latencies. Additionally, such designs revolve around custom PCBs and specialized hardware that the user needs to manufacture before being able to evaluate.

Motivated by this, we present a design that uses off-the-shelf components to provide general triggering capabilities. The system enables accurate timestamping of sensors which do not offer triggering ports (e.g. most automotive-grade LiDARs) and expands the synchronization to aligning the clocks of the \ac{mc} and \ac{obc} with standard methods, namely the IEEE 1588 \ac{ptp} (using the implementation from~\cite{schleusner2024sub}).

\begin{figure}[t!]
    \centering
    \includegraphics[width=\linewidth]{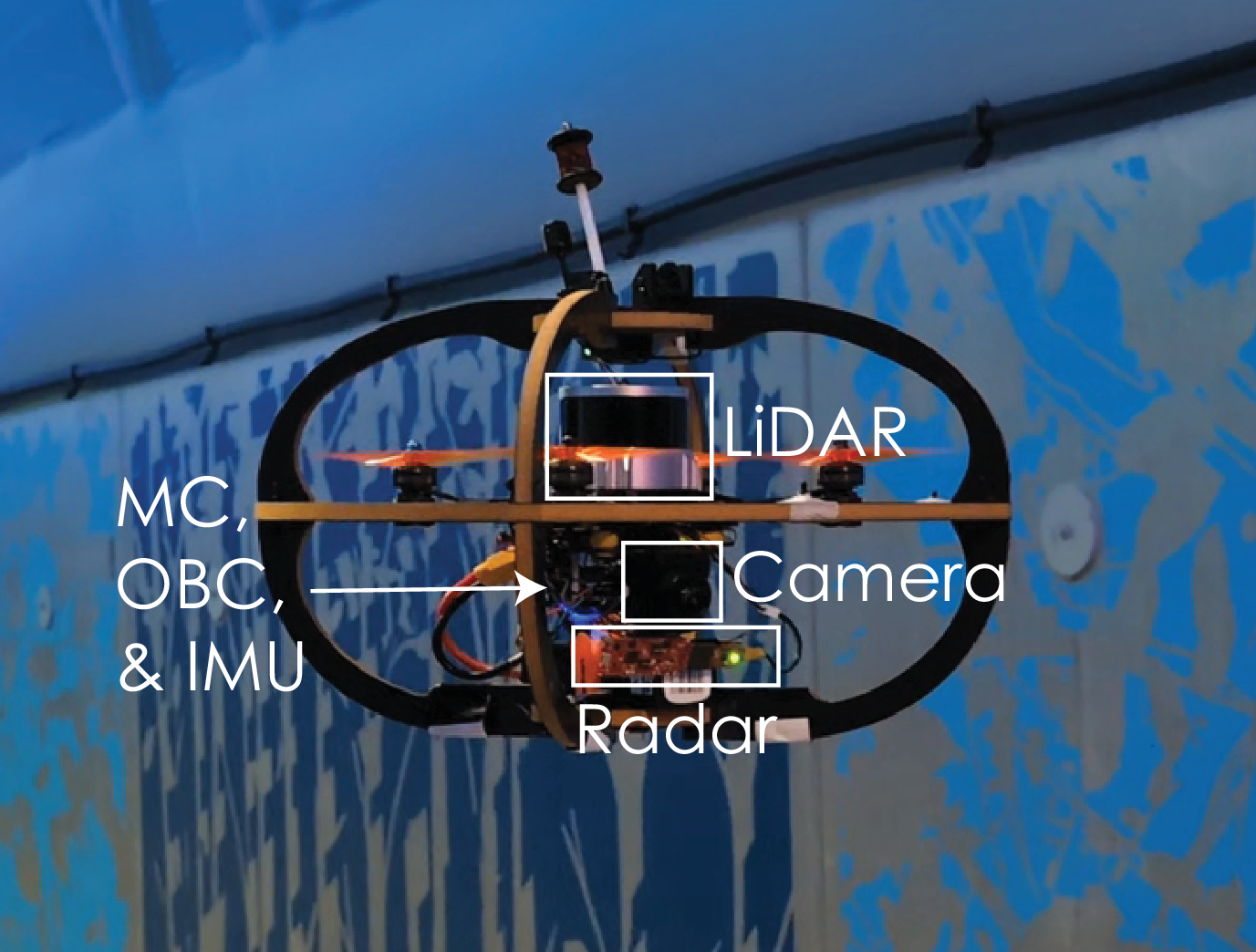}
    \vspace{-4ex}
    \caption{Robot platform which utilizes proposed system.}
    \label{fig:introduction}
    \vspace{-4ex}
\end{figure}

\section{APPROACH}\label{sec:approach}
\begin{figure}[h]
    \centering
    \includegraphics[width=0.9\linewidth]{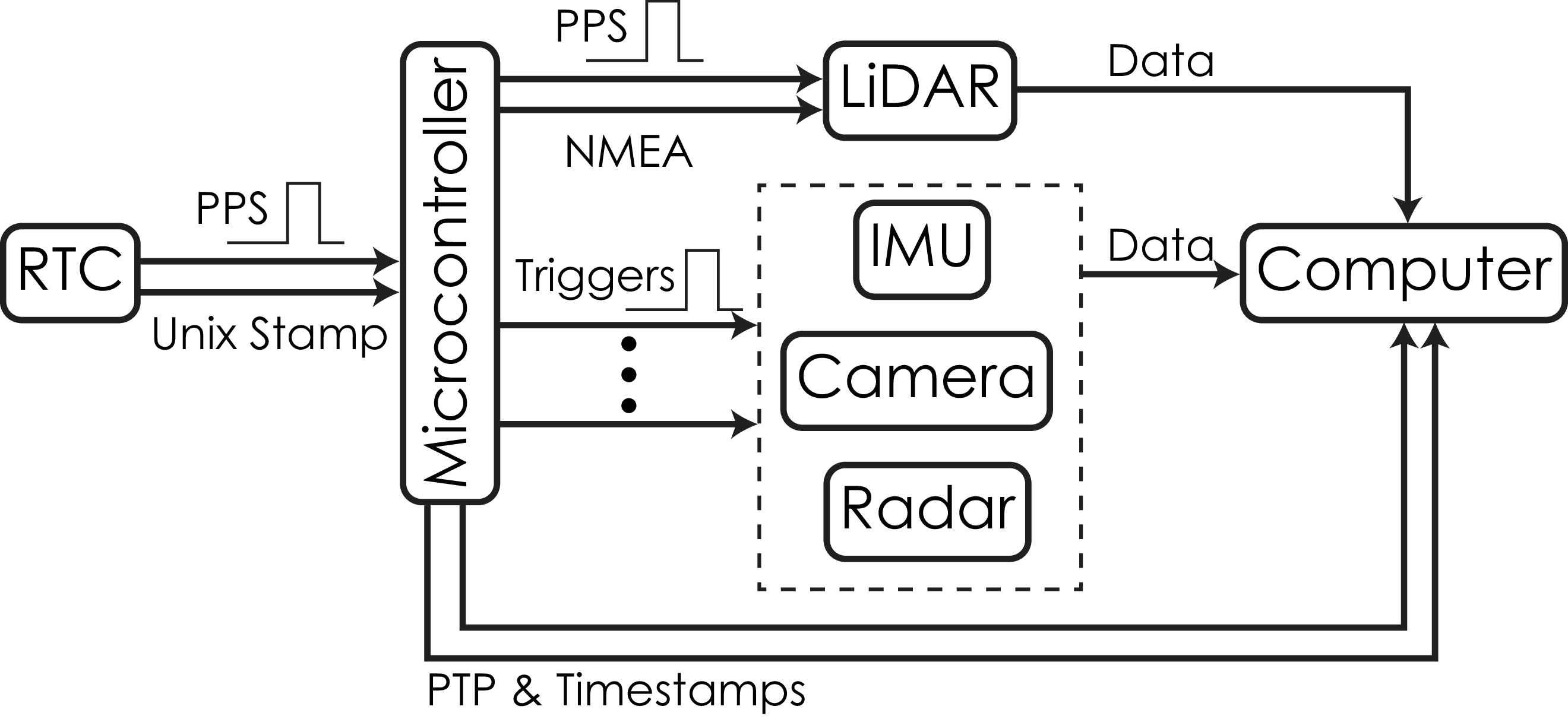}
    \vspace{-2ex}
    \caption{Proposed architecture for synchronization of \ac{rtc}-\ac{mc}-\ac{obc} system and triggering of an arbitrary set of sensors.}
    \label{fig:approach:architecture}
    \vspace{-3ex}
\end{figure}

An overview of our approach is shown in \cref{fig:approach:architecture}. We utilize a Teensy 4.1 along with a DS3231 \ac{rtc} unit. The \ac{rtc} produces a square wave with a frequency of 1Hz to which we attach an interrupt reacting to a falling edge. The event itself is timed and in the corresponding interrupt a flag is set, signaling data is ready for synchronization of the \ac{mc} to the \ac{rtc}. The captured \ac{pps} signal's timestamp is forwarded to the \ac{ptp} object, such that it can adjust the clock frequency following the output of a \ac{pi} controller acting on the calculated time offset. This establishes the synchronization between the \ac{mc} and the \ac{rtc}, resulting in a synchronized Unix time counter with resolution bounded by the \SI{24}{\mega\hertz} oscillator used by the microcontroller Ethernet interface. This behavior emulates that of a system equipped with a \ac{gnss} sensor, which provides a high-accuracy Unix time along with \ac{pps} signal, but only as long as connection persists. To enable operation in \ac{gnss}-degraded/denied environments, the \ac{rtc} is used as a less-accurate substitution that presents a near identical interaction, i.e. offering a \ac{pps} signal and the potential to communicate the corresponding Unix timestamp via serial communication.

With synchronization of the computer times in place, one can continue with the triggering and stamping of sensor measurements. To reduce computation and the possibility of preemption, the triggers are necessarily multiples of the highest frequency used (typically the IMU rate). This way, although the \ac{mc} is generally limited to few high-accuracy timers, it is able to support a much larger number of sensors, triggered at different rates with equal performance. 
This also has the advantage of reducing the potential for collision between interrupts and the resulting degradation of quality in the interrupted process.

The philosophy behind the design is that each device should be synchronized with the most accurate time source applicable. Hence, sensors that can be triggered will be triggered (camera, radar, IMU, etc), sensors that accept PPS signals along with NMEA UART messages (e.g. LiDARs) will use them, and the sensors that cannot be triggered but have an Ethernet interface 
can be synchronized via \ac{ptp} from the \ac{mc}. An example of a sensor which may not fall into any of these categories are some IMUs, which do not offer offer the possibility to be triggered. However, this is not a problem as the interrupt structure of the \ac{mc} allows for externally generated signals to initiate the triggering procedure just as well as the interrupt created by an internal high-accuracy timer. Thus, any high-rate sensor that can provide a \ac{tov} signal can still be used to drive the triggering process even if they cannot be triggered themselves.

\section{EVALUATION}\label{sec:evaluation}

The proposed system is realized using a Teensy 4.1 \ac{mc}, Khadas Vim4 \ac{obc}, and DS3231 \ac{rtc}. The system is evaluated on the basis of synchronization between compute units, as well as the triggering performance with data from a bench test.

\subsection{Synchronization}
\begin{figure}[t!]
    \centering
    \includegraphics[width=0.9\linewidth]{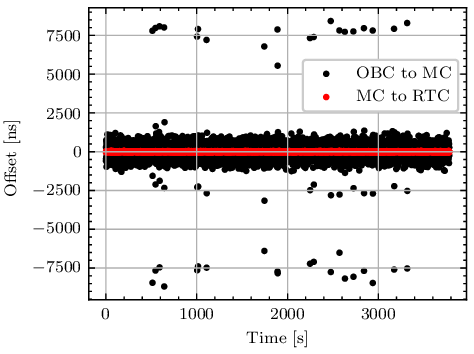}
    \vspace{-2ex}
    \caption{Offsets between \ac{rtc}-\ac{mc} and \ac{mc}-\ac{obc} during synchronization.}
    \label{fig:evaluation:synchronization}
    \vspace{-3ex}
\end{figure}

Synchronization quality between the \ac{mc} and the \ac{obc} is evaluated by examining the offset of both while the \ac{mc} is actively triggering, as seen in \cref{fig:evaluation:synchronization}. In this figure, both the synchronization performance from \ac{mc} to \ac{rtc} and from \ac{obc} to \ac{mc} can be seen. The \ac{mc} demonstrates its ability to maintain low offsets (bounded by \SI{100}{\nano\second}) with respect to the \ac{rtc} \ac{pps}, taken as pseudo \ac{gnss}. The \ac{obc} has worse performance, as a Linux machine with a non-real time operating system, but remains in the single-digit number of \si{\micro\second}. Both are several orders of magnitude less than the sampling rates of typical robotics sensors.

\subsection{Triggering}
To evaluate the triggering consistency, the proposed system is left triggering blindly for a significant period of time, i.e. the \ac{mc} is sending out signal without sensors being connected. 
From this signal, along with the reported timestamps, sampling times for the trigger signal can be calculated. The statistics are then compared in \cref{tab:evaluation:triggering} for what the authors consider to be typical sampling rates (800, 20, and \SI{10}{\hertz}). In \cref{tab:evaluation:triggering} one can see very little deviation from the expected rate, demonstrating that the system is capable of creating consistent trigger signals while simultaneously synchronizing to the \ac{rtc} and serving as a \ac{ptp} master to the \ac{obc}.

\begin{table}[h!]
    \vspace{-1ex}
    \centering
    \caption{Trigger sampling time statistics.}
    \label{tab:evaluation:triggering}
    \vspace{-2ex}
    \sisetup{
        table-format=6.3,
        table-alignment-mode=format,
        round-precision=3
    }
    \begin{tabular}{SSS}
        \toprule
        {\makecell[c]{Expected [\si{\micro\second}]}}    &{\makecell[c]{Mean [\si{\micro\second}]}}   &{\makecell[c]{Standard Deviation [\si{\micro\second}]}}\\
        \midrule
        1250   &1250.000  &0.079\\
        50000   &50000.014  &0.105\\
        100000   &100000.029  &0.119\\
        \bottomrule
    \end{tabular}
    \vspace{-2ex}
\end{table}

\section{CONCLUSIONS}\label{sec:conclusions}
A system designed to ensure high quality measurement timestamping with simultaneous synchronization between the real-time microcontroller and a more powerful onboard computer is presented. The system utilizes a microcontroller for the triggering and timestamping of 
external sensors. This timestamp information is then conveyed to a more powerful onboard computer which matches timestamps to measurements. 
The microcontroller also establishes time synchronization with the Linux host using \ac{ptp}, such that although timekeeping authority is left to the microcontroller, the onboard computer's clock is still accurate.
Furthermore, our system utilizes low-cost commercial components instead of specialized hardware while also allowing for generalizing to different sensor configurations.
Future work includes demonstration of the system design with robotic experiments.

\addtolength{\textheight}{-12cm}   %

\bibliographystyle{IEEEtran}
\bibliography{bib/general}

\end{document}